\documentclass[letterpaper]{article} % DO NOT CHANGE THIS
\usepackage{aaai25}  % DO NOT CHANGE THIS
\usepackage{times}  % DO NOT CHANGE THIS
\usepackage{helvet}  % DO NOT CHANGE THIS
\usepackage{courier}  % DO NOT CHANGE THIS
\usepackage[hyphens]{url}  % DO NOT CHANGE THIS
\usepackage{graphicx} % DO NOT CHANGE THIS
\usepackage{natbib}  % DO NOT CHANGE THIS AND DO NOT ADD ANY OPTIONS TO IT
\usepackage{caption} % DO NOT CHANGE THIS AND DO NOT ADD ANY OPTIONS TO IT
\usepackage{algorithm}
\usepackage{algorithmic}

\usepackage{newfloat}
\usepackage{listings}
\DeclareCaptionStyle{ruled}{labelfont=normalfont,labelsep=colon,strut=off} % DO NOT CHANGE THIS
\floatstyle{ruled}
\newfloat{listing}{tb}{lst}{}
\floatname{listing}{Listing}
\usepackage{graphicx}
\usepackage{multirow}
\usepackage{booktabs}
\usepackage{enumitem}
\usepackage{color}

\usepackage{float}
\usepackage{tabularx}
\usepackage{mathbbol}
\usepackage{amsmath}
\usepackage{amssymb}
\usepackage[table]{xcolor}
\usepackage{colortbl}

\newcommand{\tablestyle}[2]{\setlength{\tabcolsep}{#1}\renewcommand{\arraystretch}{#2}\centering\footnotesize}

\title{MoRe: Class Patch Attention Needs Regularization for Weakly Supervised Semantic Segmentation}
\author{
    Zhiwei Yang\textsuperscript{\rm 1,2,3},
    Yucong Meng\textsuperscript{\rm 2,3},
    Kexue Fu\textsuperscript{\rm 4},
    Shuo Wang\textsuperscript{\rm 2,3}\thanks{Corresponding authors.},
    Zhijian Song\textsuperscript{\rm 1,2,3}\footnotemark[1]
}

\affiliations{
    \textsuperscript{\rm 1}Academy for Engineering and Technology, Fudan University, Shanghai 200433, China\\
    \textsuperscript{\rm 2}Digital Medical Research Center, School of Basic Medical Sciences, Fudan University, Shanghai 200032, China\\
    \textsuperscript{\rm 3}Shanghai Key Laboratory of Medical Imaging Computing and Computer Assisted Intervention, Shanghai 200032, China\\
    \textsuperscript{\rm 4}Shandong Computer Science Center (National Supercomputer Center in Jinan)\\
    \{zwyang21, ycmeng21\}@m.fudan.edu.cn, \{fukexue, shuowang, zjsong\}@fudan.edu.cn
}

\usepackage{bibentry}
\begin{document}

\maketitle

\begin{abstract}
Weakly Supervised Semantic Segmentation (WSSS) with image-level labels typically uses Class Activation Maps (CAM) to achieve dense predictions. Recently, Vision Transformer (ViT) has provided an alternative to generate localization maps from class-patch attention. However, due to insufficient constraints on modeling such attention, we observe that the Localization Attention Maps (LAM) often struggle with the artifact issue, i.e., patch regions with minimal semantic relevance are falsely activated by class tokens. In this work, we propose MoRe to address this issue and further explore the potential of LAM. Our findings suggest that imposing additional regularization on class-patch attention is necessary. To this end, we first view the attention as a novel directed graph and propose the Graph Category Representation module to implicitly regularize the interaction among class-patch entities. It ensures that class tokens dynamically condense the related patch information and suppress unrelated artifacts at a graph level. Second, motivated by the observation that CAM from classification weights maintains smooth localization of objects, we devise the Localization-informed Regularization module to explicitly regularize the class-patch attention. It directly mines the token relations from CAM and further supervises the consistency between class and patch tokens in a learnable manner. Extensive experiments are conducted on PASCAL VOC and MS COCO, validating that MoRe effectively addresses the artifact issue and achieves state-of-the-art performance, surpassing recent single-stage and even multi-stage methods. Code is available at \textit{https://github.com/zwyang6/MoRe}.
\end{abstract}

\section{Introduction}

Weakly supervised semantic segmentation (WSSS) aims to achieve pixel-wise predictions with cheap annotations, such as points~\cite{1}, scribbles~\cite{2,3}, bounding boxes~\cite{4,5}, and image-level labels~\cite{6,7}. It significantly reduces the annotation cost of fully supervised segmentation methods~\cite{x1,x3} and has attracted increasing attention in recent years. Among all annotation forms, the image-level label is the most accessible while challenging, as it contains the least semantic information to localize objects. In this work, we focus on the WSSS paradigm with only image-level labels.

\begin{figure}[t]
    \centering
    \includegraphics[width=1.0\linewidth]{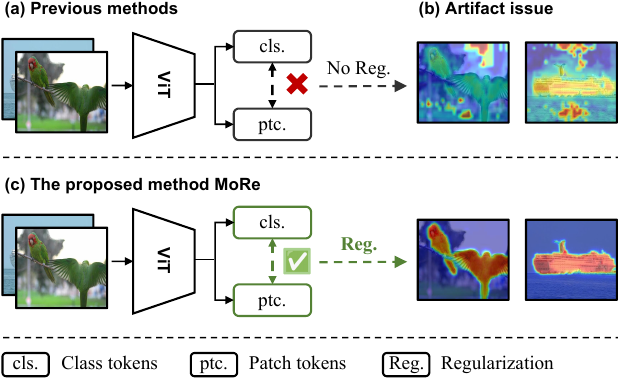}
    \caption{Our motivation. Localization Attention Maps (LAM) from ViT provide an alternative to CAM. (a) Since no regularization is conducted between class-patch attention, (b) LAM particularly suffers from the artifact issue. (c) We propose MoRe to tackle it and generate better LAM by regularizing attention among class-patch tokens.}
    \label{fig1}
    \vspace{-1em}
\end{figure}

Typically, the pipeline of WSSS can be divided into three steps. It first trains a classification network to generate Class Activation Maps (CAM)~\cite{0} with image-level labels. Then CAM is refined as pseudo labels, which are further leveraged to provide dense supervision to retrain a segmentation model~\cite{8,9}. However, due to the limited supervision, CAM intends to activate only the most discriminative parts of objects, which dramatically impairs the performance of WSSS. Recently, Vision Transformer (ViT)~\cite{10}, famous for building long-range dependency, has been widely adopted in WSSS. Benefiting from the self-attention mechanism, several studies~\cite{11,12} have shown that attention maps between class and patch tokens can reliably highlight objects, offering a promising alternative for generating precise localization maps. Inspired by it, TS-CAM~\cite{13} directly extracts the Localization Attention Maps (LAM) from class-patch attention and leverages it to generate pseudo labels. However, since the original ViT only contains a single-class token, TSCAM still struggles to perform class-specific dense predictions. To fill in this gap, MCTformer series~\cite{14,22} integrates multi-class tokens into ViT and successfully generates class-specific LAM, which demonstrates the great potential of LAM in enhancing the performance of WSSS.

However, as demonstrated in~\cite{155,16}, in order to capture long-range dependencies, ViT inclines to aggregate global semantics in low-information patches. It causes unrelated patches to be frequently correlated with class tokens during attention, leading to numerous false activations of artifact patch tokens. Previous LAM-based WSSS methods barely regularize such attention. Consequently, this issue commonly exists in ViTs (such as ViT on ImageNet~\cite{55} or DeiT~\cite{35}) and severally impairs the quality of LAM, as shown in Figure~\ref{fig1} (a, b). Importantly, unlike the notorious over-smoothness issue~\cite{18} of ViT that stems from the high similarity among patch tokens, the reported artifact issue is derived from the improper correlation between class-patch tokens. Although recent WSSS works have proposed to solve the over-smoothness~\cite{19,20}, both LAM-based and CAM-based methods particularly overlook the artifact issue from improper class-patch attention, leaving this problem unaddressed.

In this work, we find that more regularization is necessary for constructing proper class-patch attention and propose MoRe to tackle the artifact issue. First, we propose the Graph Category Representation (GCR) module to implicitly regularize the relation among class-patch tokens. GCR views the class-patch attention as a novel directed graph representation~\cite{21}. To this end, each token in ViT is regarded as a node and is further represented by the head and tail embeddings. To precisely model the correlation among class-patch entities, class-related neighbors are dynamically updated and additional edge embeddings are designed to parameterize the relation between heads and tails. Building upon this dynamic graph structure, the Graph Aggregation mechanism is designed to fully consider the knowledge from head, tail, and edge, allowing more reliable patches to aggregate class-related semantics into class tokens. 

In addition, since CAM and LAM exhibit strong alignment in object localization, we design the Localization-informed Regularization (LIR) module to explicitly regularize class-patch attention. Although previous works propose to fuse LAM with CAM~\cite{14,22}, the fusion strategy simply combines them with a multiplying operation while no optimization is conducted during attention. Instead, LIR mines token relations from CAM and further leverages the supervision to facilitate the consistency between class-patch tokens in a learnable manner. Specifically, guided by the clues from CAM, a confident relation enhancement loss is designed to promote the correlation between class and confidently related patch tokens while suppressing unrelated artifacts. With the enhanced class tokens, we also formulate an uncertain relation enhancement loss to supervise more uncertain yet relevant regions being attended by class tokens. Finally, based on the GCR and LIR modules, MoRe is seamlessly incorporated into ViT-based WSSS and effectively tackles the artifact end-to-end, producing high-fidelity LAM and pseudo labels, as shown in Figure~\ref{fig1} (c). 

The main contributions of our work are listed as follows:

\begin{itemize}
    \item This work reports and addresses the artifact issue when generating LAM from class-patch attention. We find that more regularization is needed to constrain the improper attention and propose MoRe to achieve it.
    \item Two forms of regularization are designed. We design the Graph Category Representation (GCR) module to implicitly regularize the class-patch attention, which enhances the interaction of class-related information from a graph perspective. The Localization-informed Regularization (LIR) module is proposed to explicitly constrain the relation among class-patch tokens, which facilitates the class-patch consistency in a learnable manner. 
    \item Extensive experiments are conducted on PASCAL VOC and MS COCO, demonstrating the efficacy of MoRe and the superior performance over recent single-stage methods and even sophisticated multi-stage techniques. 
\end{itemize}

\section{Related Work}

\begin{figure*}[t]
    \centering
    \includegraphics[width=1.0\linewidth]{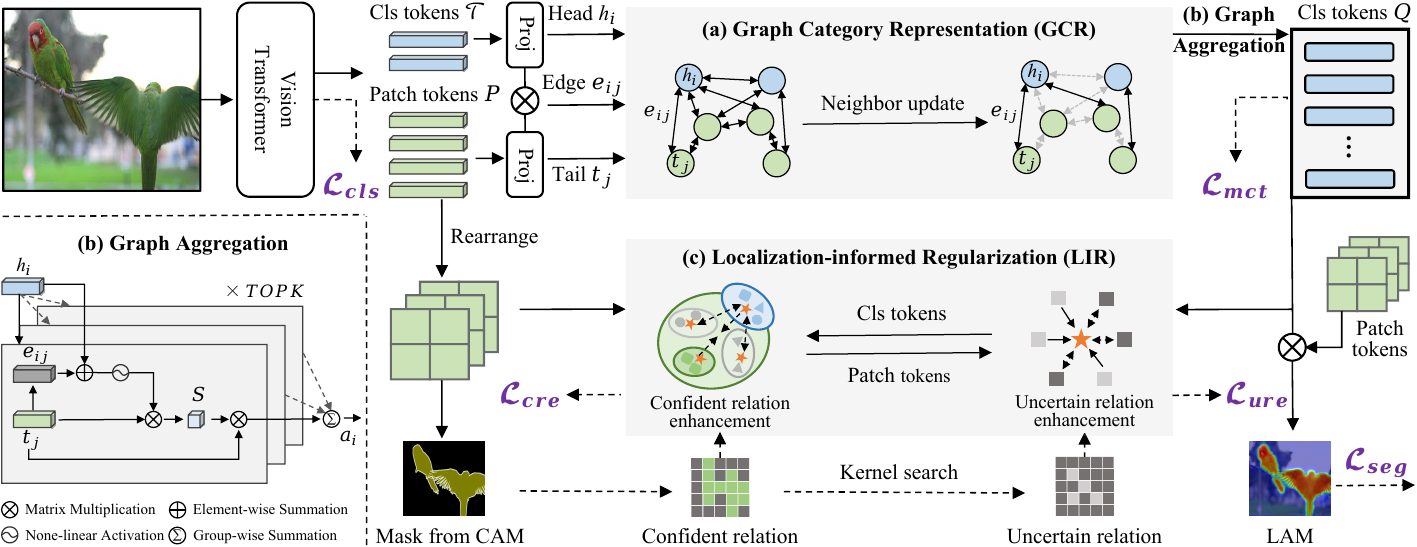}
    \caption{Overview of our MoRe. The input image is sent to ViT encoder and generates multi-class and patch tokens. (a) We first send them into Graph Category Representation (GCR) module, which takes the class-patch attention as a directed graph with the projected entities head $h_i$, tail $t_j$, and learnable edge $e_{ij}$. (b) Then Graph Aggregation mechanism is designed to condense the related tail semantics into class tokens. (c) CAM is also generated from patches. It acts as the confident and uncertain relation supervision to the proposed Localization-informed Regularization (LIR) module with two objectives $\mathcal{L}_{{cre}}$ and $\mathcal{L}_{{ure}}$. Finally, LAM is generated from the similarity score map between class-patch tokens and is used to train a segmentation decoder.}
    \label{fig2}
    \vspace{-1em}
\end{figure*}

\subsection{Weakly Supervised Semantic Segmentation}
Weakly supervised semantic segmentation (WSSS) with image-level labels particularly leverages Class Activation Maps (CAM) to achieve dense predictions. Due to the classification nature of CAM, CAM only activates the most discriminative parts of objects. To generate more complete CAM, considerable efforts have been paid with intriguing insights, such as prototype maintaining~\cite{23,24}, affinity propagation~\cite{27}, and multi-modal supervision~\cite{28}, etc. With the emergence of ViT, ViT-based WSSS has substantially mitigated this issue. AFA~\cite{27} learns affinity from attention and uses it to refine CAM. TSCAM~\cite{13} notices that class-patch attention maps in ViT can localize objects and generate more complete Localization Attention Maps (LAM). Inspired by it, MCTformer series~\cite{14,22} integrates multi-class tokens and further generates class-specific LAM, which serves as a substitute for CAM. Despite these advancements, previous LAM-based methods overlook the artifact issue. In this work, we further explore the potential of LAM and intend to address this issue by regularizing the attention between class and patch tokens.

\subsection{Graph Neural Networks}
Graph neural networks (GNN) aim to model the representation with graph elements (e.g., nodes and edges) and approximation inference~\cite{21}. With its ability to capture complex interactions within the entity topology, GNN has attracted widespread attention across various fields, such as 3D pose estimation~\cite{29} and medical image analysis~\cite{31}. For WSSS, most GNN-related works are based on convolutional neural networks (CNN). GraphNet~\cite{32} and A2GNN~\cite{33} are proposed to construct graph structures based on superpixel or affinity under the supervision of bounding boxes. GSM~\cite{34} employs GNN to capture semantic dependencies among images within the same class. Although impressive, they typically rely on pre-defined relations to construct the CNN-based graph and ignore the directed interaction among entities. In this work, we build our graph on ViT following~\cite{291,30}, while uniquely highlighting a directed topology between class-patch tokens to mine their mutual information.

\section{Methodology}

\subsection{Framework Overview}
The pipeline of MoRe is illustrated in Figure~\ref{fig2}. The input image space and classification label are defined as ${X}$ and $\mathcal{Y}=\{1,2, \ldots, C\}$, where $C$ denotes the number of categories. Given a training batch $({I}, {l})$ from image and label space, ${I} \in \mathbb{R}^{3 \times \mathcal{H} \times \mathcal{W}}$ is the training image and ${l} \in \mathcal{Y}$ is the class label. We inherit the multi-class token settings from ~\cite{14} and send $I$ into ViT, which generates the patch tokens $P \in \mathbb{R}^{L \times D}$ and multi-class tokens $\mathcal{T} \in \mathbb{R}^{C \times D}$. $L, D$ are the number of patches and the embedding dimension. Building upon this, we first send $\mathcal{T}$ and $P$ to the proposed Graph Category Representation (GCR) module, which views the class attention as a novel directed graph, shown in Figure~\ref{fig2} (a). Then Graph Aggregation mechanism is designed to fully condense the related patch semantics into class tokens, as shown in Figure~\ref{fig2} (b). With the aggregated class tokens $Q\in \mathbb{R}^{C \times D}$, the Localization-informed regularization (LIR) module is designed to further regularize the attention between class and patch tokens, as shown in Figure~\ref{fig2} (c). Finally, we generate LAM by calculating the cosine similarity score between class-patch tokens. LAM is leveraged as pseudo labels to train a decoder with segmentation loss ${L}_{{seg}}$. We also maintain a classification loss ${L}_{{cls}}$ to generate CAM, which provides localization relations for LIR module. More details are introduced in the subsequent sections.

\subsection{Graph Category Representation}
\subsubsection{Category Graph Construction} 
To suppress the artifacts from improper interaction between class-patch tokens, we view the attention as a directed graph structure and implicitly regularize the relation with learnable topological features. As shown in Figure~\ref{fig2} (a), given the multi-class tokens $\mathcal{T}$ and patches $P$ from ViT, we adopt two linear projectors to transform them into heads $H \in \mathbb{R}^{C \times D}$ and tails $T \in \mathbb{R}^{L \times D}$, where heads model the correlation to patches while tails represent the contribution from patches to heads. Since artifacts commonly exist in low-information patches, we assume that those tokens should be dynamically excluded during attention. To this end, we customize the edge embeddings $e_{ij}$ and select candidate neighbors to model the relation between head $h_{i}$ and tail $t_{j}$. We first select the top-$K$ cosine similarity scores to quantify the similarity between $h_{i}$ and its most related patches, which are formulated as:
\begin{equation}
    r_{i} = softmax(TOPK\{h_{i}^{T}t_{j}\}_{j=1}^{L}) .
    \label{eq:1}
\end{equation}

To exclude the unrelated artifact tokens, we propose updating neighbors with the tail candidates ranked by the similarity score $r_i$. The index of most related top-$K$ tail neighbors to head $h_{i}$ can be denoted as follows: 
\begin{equation}
N_{i}=\{j \mid j \in \operatorname{argmax}_{T O P K}\{h_{i}^{T} t_{j}\}_{j=1}^{L}\}.
    \label{eq:2}
\end{equation}

Based on the updated neighbors, the edge embeddings maintained between head $h_{i}$ and tail $t_{j}$ are formulated as:
\begin{equation}
e_{i j}=r_{i j} t_{j}+\left(1-r_{i j}\right) h_{i}, j \in N_{i} .
    \label{eq:3}
\end{equation}

With the generated entities, we delineate the attention between class and patch tokens as a dynamic directed graph $G=\{V,R,Z,E\}$, where $V$ is the nodes corresponding to class and patch tokens, $Z$ represents the head and tail. $E$ denotes the edge. $R=\{(h,e,t):(h,t) \in Z, e \in E\}$ generalizes head, tail, and edge, representing the directed information on the directed edge. Compared to the vanilla class attention, the topological graph structure effectively reasons the relation among class and patch tokens by considering the triplets of head, tail, and the learnable directed edge.

\subsubsection{Graph Aggregation Mechanism} 
Building upon the constructed directed graph $G$, we further propose the Graph Aggregation mechanism to aggregate reliable knowledge from patches into class tokens, as shown in Figure~\ref{fig2} (b). To model the different importance of knowledge propagated from tail to head, a weighting factor $S(h_{i},e_{ij},t_{j})$ is designed to guarantee the propagation process, which is determined by the triplets of head, tail, and edge as:
\begin{equation}
S(h_{i}, e_{i j}, t_{j})=\operatorname{softmax}(t_{j}^{T} \sigma(h_{i}+e_{i j})),
    \label{eq:4}
\end{equation}
where $\sigma(\cdot)$ is the gated activation function $tanh(\cdot)$. 

Then the knowledge from related neighbors to $h_{i}$ is denoted as: $a_i=\sum_{j \in N_i} S\left(h_i, e_{i j}, t_j\right) t_j$. By representing relationships among triplets as edge-based knowledge, head nodes can effectively receive and capture signals from tail nodes. Finally, we generate the graph-regularized class tokens $Q$ by fusing the knowledge with the initial heads. We introduce a bi-directional aggregating strategy to facilitate the information propagation:
\begin{equation}
Q=\delta_1\left(w_1\left(h_i+a_i\right)\right)+\delta_2\left(w_2\left(a_i \odot h_i\right)\right),
    \label{eq:5}
\end{equation}
where $\delta_1/\delta_2$ adopt LeakyReLU, $w_1/w_2$ are project matrix, and $\odot$ is the element-wise multiplication.

\subsection{Localization-informed Regularization}
\subsubsection{Confident Relation Enhancement} Since CAM holds a high alignment of localization objects with LAM, we further leverage such prior to generate confident relation $M_c$ and uncertain relation $M_u$. They explicitly regularize the class-patch attention in a learnable manner, as shown in Figure~\ref{fig2} (c). Formally, we leverage $\mathcal{L}_{cls}$ to maintain a classification head to generate CAM $\in \mathbb{R}^{n_h \times n_w \times C}$ by projecting the classification matrix on rearranged $P$, where $n_h \times n_w$ is the feature size. We adopt a multi-threshold filtering strategy to refine CAM into reliable masks, which contain foreground, background, and uncertain regions:
\begin{equation}
    M_a=\left\{\begin{array}{ll}
    \operatorname{argmax}(C A M_{i, j,:}), &\text{if max }(C A M_{i, j,:})>\lambda_h, \\
    0, &\text{if max }(CAM_{i, j,:})<\lambda_l, \\
    255, &\text { otherwise},
    \end{array}\right.
    \label{eq:6}
\end{equation}
where $M_a\in \mathbb{R}^{n_h \times n_w}$, thresholds $0<\lambda_l<\lambda_h<1$, 0 and 255 denote the index of background and uncertain regions.

To promote the correlation between class and related patch tokens while suppressing unrelated artifacts, we extract the reliable foreground regions from $M_{a}$ and take it as the confident relation map $M_c\in \mathbb{R}^{n_h \times n_w}$ to guide the class-patch attention. Specifically, for patch token $p_{ij}$ in $P$, we calculate the cosine similarity with class tokens $Q$. Then we leverage the class index on $M_{c}$ as the relation supervision. If the class token $q_{l} \in Q$ shows the same class index $l$ with the pixel $(i,j)$ on $M_{c}$, we view the correlation between class token $q_{l}$ and patch token $p_{ij}$ as positive while those with different classes are negative pairs. The contrast between class-patch tokens is achieved by:
\begin{equation}
\mathcal{L}_{cre}=-\frac{1}{N_c^{+}} \sum_{q_l \in Q_l} \sum_{p_{{ij}}^{+} \in P_{l}^{+}} \log \frac{\exp (q_l^{{T}} {p}_{{ij}}^{+} / \tau)}{\sum_{{p}_{{ij}} \in {P}} \exp \left({q}_l^{{T}} {p}_{{ij}} / \tau\right)},
    \label{eq:7}
\end{equation}
where $N_c^{+}$ denotes the number of positive pairs between class tokens and patch tokens. $Q_l$ is the class token set with class label $l$, $P_{l}^{+}$ is the patch token set that shows the same class index $l$ with class token $q_{l}$. $\tau$ is the temperature factor to control the sharpness of contrast. 

It is noted that the contrast benefits from two folds: 1) it explicitly improves the correlation between class and related patch tokens, which helps reliably activate patches and suppress the artifacts. 2) it directly promotes discrimination of class tokens, which further reduces the overlapping attention regions derived from ambiguity among class tokens.

\subsubsection{Uncertain Relation Enhancement}
Due to the incompleteness nature of CAM, CAM-informed LAM still struggles with the issue that only most deterministic patches are activated by class tokens. To alleviate this problem, we further extract uncertain relation map $M_u\in \mathbb{R}^{n_h \times n_w}$ from $M_{a}$ and leverage it to supervise more uncertain yet relevant regions being activated by class tokens. Since $M_{u}$ is inevitably noisy, we design an uncertain mining kernel to search comparatively reliable patches. Specifically, given a $d \times d$ kernel centered at position $(i,j)$, we send it to walk through both $M_{u}$ and $M_{c}$. Only the uncertain patch token $p_{ij}$ is selected when the number of both foreground and uncertain pixels in the kernel exceeds a certain proportion $\varphi$. The search process for uncertain yet relevant patch tokens is formulated as:
\begin{equation}
U=\left\{u_{i j} \in P_u: \operatorname{kernel}\left(i j, M_c, M_u\right) / (d \times d)>\varphi\right\},
    \label{eq:8}
\end{equation}
where $P_u$ is the uncertain patch candidates whose class index on $M_{a}$ is 255. ${kernel}(\cdot)$ is the kernel searching operator, and $\varphi$ is the proportion value to select patch tokens.

With the enhanced class tokens $Q$ and the selected uncertain patch tokens $U$, we make the class tokens act as class centroids and pull uncertain patch tokens correlated to them. Formally, given the input image with class label $l$, we view the class token $q_{l}$ and the selected uncertain patch token $u_{ij}$ as positive pairs and maximize the similarity between them. For class token $q_{-l}$ whose class index is not $l$, we view the relation with $u_{ij}$ as negative pairs and minimize the similarity. The process can be expressed as:
\begin{equation}
\mathcal{L}_{{ure}}=\frac{1}{{N}_{{u}}^{+}} \sum_{}(1-\operatorname{cosim}({q}_l^{{T}} u_{i j}))+\frac{1}{{N}_{{u}}^{-}} \sum_{} \operatorname{cosim}({q}_{-l}^{{T}} u_{i j}),
    \label{eq:9}
\end{equation}
where $q_{l} \in {Q}_l, q_{-l} \in {Q}_{-l}, u_{i j} \in U$, ${N}_{{u}}^{+}/{N}_{{u}}^{-}$ denotes the number of positive/negative pairs, respectively. $cosim(\cdot)$ denotes the computation of cosine similarity.

\subsection{Training Objectives}
\label{sec3.3}
As shown in Figure~\ref{fig2}, MoRe consists of two class-patch attention regularization losses, i.e., $\mathcal{L}_{{cre}}$ and $\mathcal{L}_{{ure}}$, and a multi-label soft margin classification loss $\mathcal{L}_{{cls}}$. We incorporate $\mathcal{L}_{{mct}}$ to further promote the discrepancy of class tokens following~\cite{22}. Denoting the weight factors of $\alpha$ and $\beta$, the optimizing objective of MoRe is formulated as:
\begin{equation}
{L}_{ {MoRe}}={L}_{ {cls}}+{L}_{ {mct}}+\alpha {L}_{cre}+\beta {L}_{ure}.
    \label{eq:10}
\end{equation}

In addition, MoRe trains segmentation decoder end-to-end. The loss ${L}_{seg}$ for segmentation uses cross-entropy. Therefore, the overall loss for MoRe is: ${L}={L}_{ {MoRe}}+\gamma {L}_{ {seg }}$. Following the prevalent settings in single-stage WSSS~\cite{19,51}, regularization losses to improve CAM and segmentation masks are also adopted. 

\section{Experiments}
\begin{table}[t]
    \centering
    \small
    \setlength{\tabcolsep}{1mm}
    \begin{tabular}{l|c|c|cc}
        \toprule
        Method & Sup. & Net. & Seed & Mask \\
        \midrule
        \multicolumn{5}{l}{\textbf{\textit{Multi-stage WSSS methods}}.} \\
        RIB~\cite{36}     & $\mathcal{I}+\mathcal{S}$ & RN101 & 56.5 & 70.6 \\
        L2G~\cite{38}     & $\mathcal{I}+\mathcal{S}$ & RN101 & - & 71.9 \\
        $\dagger$CLIMS~\cite{53}                & $\mathcal{I}+\mathcal{L}$ & RN38  & 56.6 & 70.5 \\
        $\dagger$CLIP-ES~\cite{28} & $\mathcal{I}+\mathcal{L}$ & RN101  & 70.8 & 75.0 \\
        $\dagger$CPAL~\cite{24}  & $\mathcal{I}+\mathcal{L}$ & RN101  & 71.9 & 75.8 \\
        $\dagger$PSDPM~\cite{47}  & $\mathcal{I}+\mathcal{L}$ & RN101  & - & 77.3 \\
        CDA~\cite{40}      & $\mathcal{I}$ & RN101 & 58.4 & 66.4 \\
        ReCAM~\cite{41}      & $\mathcal{I}$ & RN101 & 54.8 & 70.0 \\
        $\dagger$MCTformer~\cite{14}           & $\mathcal{I}$ & RN38  & 61.7 & 69.1 \\
        $\dagger$LPCAM~\cite{52}           & $\mathcal{I}$ & RN50 & 65.3 & 72.7 \\
        $\dagger$MCTformer+~\cite{22}  & $\mathcal{I}$ & RN101  & 68.8 & 76.2 \\
        SFC~\cite{43}  & $\mathcal{I}$ & RN101  & 64.7 & 73.7 \\
        $\dagger$CTI~\cite{48} & $\mathcal{I}$ & RN101  & 69.5 & 73.7 \\
        \midrule
        \multicolumn{5}{l}{\textbf{\textit{Single-stage WSSS methods}}.} \\
        1Stage~\cite{49}   & $\mathcal{I}$ & RN38  & - & 66.9 \\
        $\dagger$AFA~\cite{27}  & $\mathcal{I}$ & MiT-B1& 65.0 & 68.7 \\
        $\dagger$ViT-PCM~\cite{50}         & $\mathcal{I}$ & ViT-B & 67.7 & 71.4 \\
        $\dagger$ToCo~\cite{19}    & $\mathcal{I}$ & ViT-B & 71.6 & 72.2 \\
        $\dagger$DuPL~\cite{51}    & $\mathcal{I}$ & ViT-B & - & 75.1 \\
        $\dagger$\cellcolor[HTML]{EFEFEF}\textbf{MoRe-CAM(Ours)}        & $\cellcolor[HTML]{EFEFEF}\mathcal{I}$ & \cellcolor[HTML]{EFEFEF}ViT-B & \cellcolor[HTML]{EFEFEF}\textbf{76.9} & \cellcolor[HTML]{EFEFEF}\textbf{79.7} \\
        $\dagger$\cellcolor[HTML]{EFEFEF}\textbf{MoRe-LAM(Ours)}        & $\cellcolor[HTML]{EFEFEF}\mathcal{I}$ & \cellcolor[HTML]{EFEFEF}ViT-B & \cellcolor[HTML]{EFEFEF}\textbf{77.0} & \cellcolor[HTML]{EFEFEF}\textbf{80.0} \\
        \bottomrule
    \end{tabular}
    \vspace{-.5em}
    \caption{Performance comparison of pseudo labels on PASCAL VOC train set. Sup. denotes the supervision type. $\mathcal{I}$: image-level labels. $\mathcal{S}$: saliency maps. $\mathcal{L}$: Language. $\dagger$: methods based on Vision Transformer.}
    \label{tab1}
    \vspace{-1.em}
\end{table}

\begin{table}[t]
    \centering
    \small
    \tablestyle{4.7pt}{0.9789}
    \setlength{\tabcolsep}{1.mm}
    \begin{tabular}{lcccc}
    \toprule
    \multicolumn{1}{l|}{\multirow{2}{*}{Method}} & \multicolumn{1}{c|}{\multirow{2}{*}{Net.}} & \multicolumn{2}{c|}{VOC}                           & COCO          \\ \cmidrule{3-5} 
    \multicolumn{1}{l|}{}                        & \multicolumn{1}{c|}{}                      & Val           & \multicolumn{1}{c|}{Test}          & Val           \\ \midrule
    \multicolumn{5}{l}{\textit{\textbf{Multi-stage WSSS methods.}}}                                                                                                \\
    \multicolumn{1}{l|}{RIB~\cite{36}}                     & \multicolumn{1}{c|}{RN101}                 & 70.2          & \multicolumn{1}{c|}{70.0}          & 43.8          \\
    \multicolumn{1}{l|}{L2G~\cite{38}  }                     & \multicolumn{1}{c|}{RN101}                 & 72.1          & \multicolumn{1}{c|}{71.7}          & 44.2          \\
    \multicolumn{1}{l|}{RCA~\cite{26}}                     & \multicolumn{1}{c|}{RN38}                  & 72.2          & \multicolumn{1}{c|}{72.8}          & 36.8          \\
    \multicolumn{1}{l|}{CDA~\cite{40}}                     & \multicolumn{1}{c|}{RN38}                  & 66.1          & \multicolumn{1}{c|}{66.8}          & 33.2          \\
    \multicolumn{1}{l|}{ESOL~\cite{42}}                    & \multicolumn{1}{c|}{RN101}                 & 69.9          & \multicolumn{1}{c|}{69.3}          & 42.6          \\
    \multicolumn{1}{l|}{$\dagger$MCTformer~\cite{14}}               & \multicolumn{1}{c|}{RN38}                  & 71.9          & \multicolumn{1}{c|}{71.6}          & 42.0          \\
    \multicolumn{1}{l|}{$\dagger$CLIP-ES~\cite{28}}                 & \multicolumn{1}{c|}{RN101}                 & 72.2          & \multicolumn{1}{c|}{72.8}          & 45.4          \\
    \multicolumn{1}{l|}{$\dagger$OCR~\cite{44}}                     & \multicolumn{1}{c|}{RN38}                  & 72.7          & \multicolumn{1}{c|}{72.0}          & 42.5          \\
    \multicolumn{1}{l|}{$\dagger$BECO~\cite{45}}                    & \multicolumn{1}{c|}{RN101}                 & 73.7          & \multicolumn{1}{c|}{73.5}          & 45.1          \\
    \multicolumn{1}{l|}{$\dagger$MCTformer+~\cite{22}}              & \multicolumn{1}{c|}{RN38}                  & 74.0          & \multicolumn{1}{c|}{73.6}          & 45.2          \\
    \multicolumn{1}{l|}{$\dagger$CTI~\cite{48}}                     & \multicolumn{1}{c|}{RN101}                 & 74.1          & \multicolumn{1}{c|}{73.2}          & 45.4          \\
    \multicolumn{1}{l|}{$\dagger$CPAL~\cite{24}}                    & \multicolumn{1}{c|}{RN101}                 & 74.5          & \multicolumn{1}{c|}{74.7}          & 46.3          \\ \midrule
    \multicolumn{5}{l}{\textit{\textbf{Single-stage WSSS methods.}}}                                                                                               \\
    \multicolumn{1}{l|}{1Stage~\cite{49}}                  & \multicolumn{1}{c|}{RN38}                  & 62.7          & \multicolumn{1}{c|}{64.3}          & -             \\
    \multicolumn{1}{l|}{SLRNet~\cite{27}}                  & \multicolumn{1}{c|}{RN38}                  & 67.2          & \multicolumn{1}{c|}{67.6}          & 35.0          \\
    \multicolumn{1}{l|}{$\dagger$AFA~\cite{27}}                     & \multicolumn{1}{c|}{MiT-B1}                & 66.0          & \multicolumn{1}{c|}{66.3}          & 38.9          \\
    \multicolumn{1}{l|}{$\dagger$ViT-PCM~\cite{50}}                 & \multicolumn{1}{c|}{ViT-B}                 & 70.3          & \multicolumn{1}{c|}{70.9}          & -             \\
    \multicolumn{1}{l|}{$\dagger$ToCo~\cite{19}}                    & \multicolumn{1}{c|}{ViT-B}                 & 71.1          & \multicolumn{1}{c|}{72.2}          & 42.3          \\
    \multicolumn{1}{l|}{$\dagger$DuPL~\cite{51} }                    & \multicolumn{1}{c|}{ViT-B}                 & 73.3          & \multicolumn{1}{c|}{72.8}          & 44.6          \\
    \multicolumn{1}{l|}{\cellcolor[HTML]{EFEFEF}\textbf{$\dagger$MoRe(Ours)}} & \multicolumn{1}{c|}{\cellcolor[HTML]{EFEFEF}\textbf{ViT-B}} & \cellcolor[HTML]{EFEFEF}\textbf{76.4} & \multicolumn{1}{c|}{\cellcolor[HTML]{EFEFEF}\textbf{75.0}} & \cellcolor[HTML]{EFEFEF}\textbf{47.4}             \\ \bottomrule
    \end{tabular}
    \vspace{-.5em}
    \caption{Semantic segmentation results on PASCAL VOC and MS COCO in terms of mIoU(\%). Net. denotes the backbone for single-stage methods or segmentation network for the multi-stage. $\dagger$: methods based on Vision Transformer.}
    \label{tab2}
    \vspace{-1.em}
\end{table}

\subsection{Experimental Settings}

\subsubsection{Datasets and Evaluation Metrics}
The proposed method is evaluated on PASCAL VOC 2012~\cite{56} and MS COCO 2014~\cite{57}. PASCAL VOC contains 21 classes. Following~\cite{27,19}, the augmented data with $10,582$ images are used for training, $1,449$ for validating, and $1,456$ for testing. MS COCO includes 81 classes. $82,081$ and $40,137$ images are used for training and validation. Mean Intersection-over-Union (mIoU) is the main evaluation metric. Following~\cite{7}, the confusion ratio is also adopted to validate the efficacy of suppressing false positives from artifact issues.

\subsubsection{Implementation Details}
MoRe adopts ViT-B/16 pretrained on ImageNet~\cite{54} as encoder. The decoder in this work uses DeepLab-LargeFOV~\cite{58}. Following the training settings~\cite{51,22,x5}, the AdamW optimizer with an initial learning rate $6e-5$ is used. The training images are augmented with random resized cropping into $448 \times 448$, random horizontal flipping, and color jittering. We maintain neighbors with $K=392$ tail nodes. For kernel searching, we set uncertain pixels in $M_u$ as $1$ and foregrounds in $M_c$ as $2$. When the sum of both masks in kernel exceeds $\varphi=1.2$ times the kernel size $d\times d$, the uncertain patch token is selected. The loss weight factors $(\alpha, \beta, \gamma)$ for our training objectives are $(0.2, 0.1, 0.12)$. All experiments are conducted on RTX 3090. More details can be found in \textit{Appendix}.
\begin{figure*}[t!]
    \centering
    \includegraphics[width=1.0\linewidth]{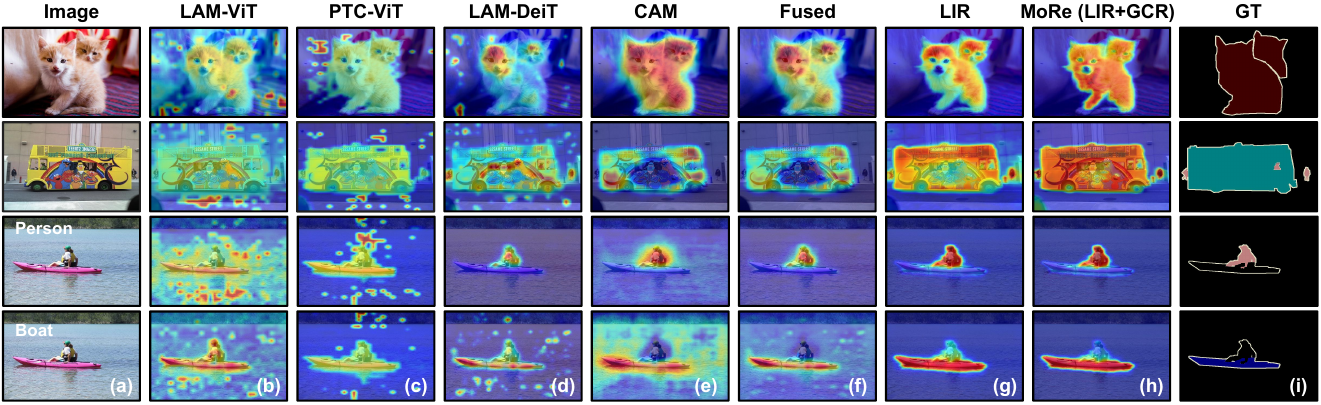}
    \caption{Visualization of LAM. (a) image. (b) LAM on ViT pretrained on ImageNet. (c) LAM with PTC loss~\cite{19} for tackling over-smoothness of ViT. (d) LAM on DeiT. (e) Patch CAM from MCTformer+~\cite{22}. (f) Refined LAM with fusion strategy by multiplying both LAM and CAM. (g) LAM with our designed LIR module. (h) LAM with both our designed LIR and GCR modules. (i) Ground truth. More visualized results are showcased in \textit{Appendix}.}
    \label{fig3}
    \vspace{-1.3em}
\end{figure*} 

\begin{figure}[t]
    \centering
    \includegraphics[width=1.0\linewidth]{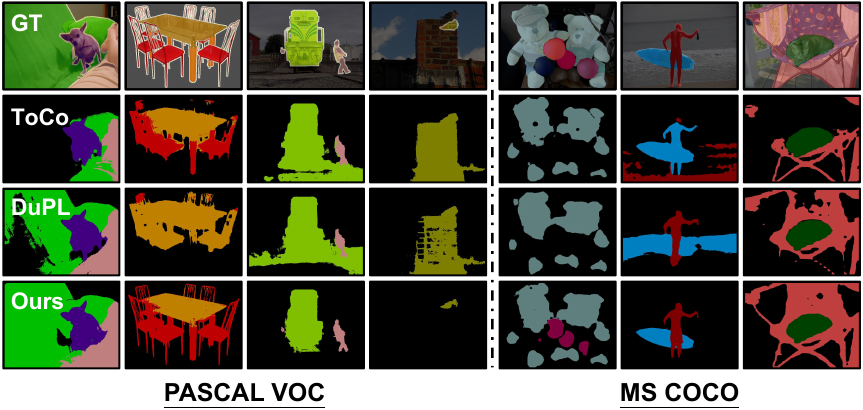}
    \vspace{-1.5em}
    \caption{Segmentation visualization with SOTA single-stage methods (i.e., ToCo and DuPL) on VOC and COCO. }    
    \label{fig4}
    \vspace{-1.3em}
\end{figure}

\subsection{Comparison with State-of-the-arts}
\paragraph{Evaluation of Pseudo Masks}
The quantitative comparisons of localization seeds and pseudo masks are reported in Table~\ref{tab1}. Without post-processing, MoRe generates seeds with $77.0\%$ mIoU for LAM and $76.9\%$ for CAM. Both seeds surpass most multi-stage methods with sophisticated refinements. With DenseCRF~\cite{59}, our pseudo masks from LAM and CAM are further improved to $80.0\%$ and $79.7\%$, respectively. With regularization on class-patch attention, MoRe shows the superiority of $3.8\%$ over the LAM-based counterpart~\cite{22} and $4.9\%$ over single-stage SOTA~\cite{51}. The qualitative comparisons are showcased in Figure~\ref{fig3}, which shows the competence of MoRe to generate better LAM as well.

\paragraph{Performance of Segmentation}
Table~\ref{tab2} reports the segmentation comparisons of MoRe with other SOTA methods on VOC and COCO. MoRe achieves $76.4\%$ mIoU on VOC val set, which noticeably outperforms both single-stage and even sophisticated multi-stage methods by at least $3.1\%$ and $1.9\%$. It achieves $75.0\%$ on VOC test and $47.4\%$ on COCO val set, which also gains improvement over other methods. 

Figure~\ref{fig4} visualizes segmentation results on VOC and COCO. It shows that MoRe clearly differentiates categories and generates complete predictions with precise boundaries. For example, MoRe successfully segments chairs from a table and even captures the outline of chair arms, while the recent methods confuse the objects (case in column $2$).

\subsection{Ablation Studies and Analysis}
\subsubsection{Efficacy of Key Components}
Quantitative ablation results are detailed in Table~\ref{tab.3}. We leverage ViT/B pretrained on ImageNet as the baseline, which also inherits multi-class tokens supervised by objectives from~\cite{14} end-to-end. As reported, it cannot achieve satisfying predictions. The proposed $\mathcal{L}_{cre}$ in LIR module explicitly promotes the correlation between class and related patch tokens. $\mathcal{L}_{ure}$ reliably searches uncertain yet relevant patch tokens and ensures them being activated. Without supervision from $\mathcal{L}_{cre}$, the precision and mIoU heavily drop by $9.2\%$ and $11.4\%$, which explains that $\mathcal{L}_{cre}$ not only suppresses artifacts, but also reduces the over-lapping attention region among class tokens by improving the discrimination. Without supervision from $\mathcal{L}_{ure}$, the recall result also drops from $88.0\%$ to $85.7\%$. The proposed GCR module dynamically updates related neighbors with learnable edge information and controls the information propagated to class tokens. Without the graph structure GCR, the segmentation performance heavily decreases by $4.3\%$. The above results confirm that the proposed components effectively reduce artifacts and contribute to generating high-quality semantic predictions.

\begin{table}[tbp]
\centering
    \tablestyle{3.4pt}{1}
    \scalebox{.99}
    {
    \footnotesize
    \begin{tabularx}{\linewidth}{@{}l|cccccc@{}}
    \toprule
    Conditions       & \multicolumn{1}{l}{$\mathcal{L}_{cre}$} & $\mathcal{L}_{ure}$ & \multicolumn{1}{l}{GCR} & Precision        & Recall     & \multicolumn{1}{l}{mIoU} \\ \midrule
    Baseline (ViT-B) &                         &     &                              & 76.2             & 46.5             & 41.6                     \\
    w/o $\mathcal{L}_{cre}$          &                         & \pmb{$\checkmark$}   & \pmb{$\checkmark$}                            & 75.3          & 80.1          & 65.0                     \\
    w/o $\mathcal{L}_{ure}$          & \pmb{$\checkmark$}                       &     &\pmb{$\checkmark$}                              & 84.3          & 85.7          & 74.4                     \\
    w/o GCR     & \pmb{$\checkmark$}                       & \pmb{$\checkmark$}   &                              & 81.6          & 85.5          & 72.1                     \\
    
    \rowcolor[HTML]{EFEFEF} 
    {\color[HTML]{000000} \textbf{MoRe}}         & {{\color[HTML]{000000} \pmb{$\checkmark$}}}        & {{\color[HTML]{000000} \pmb{$\checkmark$}}} &{{\color[HTML]{000000} \pmb{$\checkmark$}}}   &{\color[HTML]{000000} \textbf{84.5}} &{\color[HTML]{000000} \textbf{88.0}} &{\color[HTML]{000000} \textbf{76.4}}\\   \bottomrule
    \end{tabularx}
    }
   \vspace{-0.5em}
    \caption{Ablation study of MoRe on VOC val set.}
   \vspace{-1em}
   \label{tab.3}
\end{table}

\begin{table}[t]
    \centering
    % \vspace{-1em}
    \tablestyle{4.7pt}{1.1}
    \scalebox{0.99}
    {
    \footnotesize
    \begin{tabularx}{\linewidth}{@{}l|cccc@{}}
    \toprule
                       &{ToCo}          & {DuPL}        & \multicolumn{2}{c}{\color[HTML]{000000}\textbf {MoRe(Ours)}} \\ \midrule
    Aeroplane & 80.6 {(0.19)}             & 77.9 {(0.26)}              & 85.2$_{\textbf{\textcolor{teal}{+4.6}}}$  &\cellcolor[HTML]{EFEFEF}{\color[HTML]{000000} \textbf{{(0.13)}}}  $_{\textbf{\textcolor{teal}{-7.0\%}}}$            \\
    Bird     & 68.4 {(0.42})             & 81.7 {(0.20)}              & 91.7$_{\textbf{\textcolor{teal}{+10.0}}}$ &\cellcolor[HTML]{EFEFEF}{\color[HTML]{000000} \textbf{{(0.05)}}}$_{\textbf{\textcolor{teal}{-15.0\%}}}$              \\
    Boat  & 45.4 {(1.11)}             & 58.7 {(0.53)}              & 73.0$_{\textbf{\textcolor{teal}{+14.3}}}$ &\cellcolor[HTML]{EFEFEF}{\color[HTML]{000000} {\textbf{(0.23)}}}$_{\textbf{\textcolor{teal}{-30.0\%}}}$              \\
    TV monitor    & 63.1 {(0.44)}              & 45.5 {(1.10)}              & 64.3$_{\textbf{\textcolor{teal}{+1.2}}}$ &\cellcolor[HTML]{EFEFEF}{\color[HTML]{000000} \textbf{(0.38)}}$_{\textbf{\textcolor{teal}{-6.0\%}}}$             \\
    Car    & 83.3 {(0.11)}              & 77.5 {(0.20)}              & 83.7$_{\textbf{\textcolor{teal}{+0.4}}}$ &\cellcolor[HTML]{EFEFEF}{\color[HTML]{000000} \textbf{(0.10)}}$_{\textbf{\textcolor{teal}{-1.0\%}}}$              \\
    Sofa   & 43.8 {(0.77)}             & 53.9 {(0.59)}              & 55.1$_{\textbf{\textcolor{teal}{+1.2}}}$ &\cellcolor[HTML]{EFEFEF}{\color[HTML]{000000} {\textbf{(0.44)}}}$_{\textbf{\textcolor{teal}{-15.0\%}}}$              \\
    Potted plant    &  56.5 {(0.59)}             & 60.7 {(0.54)}              & 67.7$_{\textbf{\textcolor{teal}{+7.0}}}$ &\cellcolor[HTML]{EFEFEF}{\color[HTML]{000000} { \textbf{(0.20)}}}$_{\textbf{\textcolor{teal}{-34.0\%}}}$              \\ \midrule
   {Average} & {71.1 {(0.32)}} & {73.3 {(0.31)}} 
   & 76.4$_{\textbf{\textcolor{teal}{+3.1}}}$ &\cellcolor[HTML]{EFEFEF}{{\color[HTML]{000000} {\textbf{(0.22)}}}}$_{\textbf{\textcolor{teal}{-9.0\%}}}$ \\ \bottomrule
    \end{tabularx}
    }
   \vspace{-.5em}
    
    \caption{Class-specific performance of IoU and confusion ratio (in bracket) with recent methods on VOC val set.}
   \label{tab.4}
   \vspace{-1.3em}
\end{table}

\paragraph{Effectiveness of Suppressing Artifacts}
Qualitative ablation results are further illustrated in Figure~\ref{fig3} to investigate the impact of proposed modules. As shown in Figure~\ref{fig3} (b), the ViT baseline typically suffers from the artifact issue and cannot achieve reasonable localization. The proposed LIR module directly maximizes the similarity between class and related patch tokens while minimizing the irrelevant. As seen from Figure~\ref{fig3} (g), LIR significantly suppresses the artifacts and activates more non-deterministic regions, which verifies the efficacy of $\mathcal{L}_{cre}$ and $\mathcal{L}_{ure}$, respectively. The proposed GCR module works by dynamically constructing related neighbors and precisely aggregating useful semantics into class tokens from a graph perspective. Figure~\ref{fig3} (h) demonstrates that MoRe models robust representation of categories and activates objects with higher confidence. It validates that GCR further improves the quality of LAM. 

\begin{table}[t!]
\setlength{\abovecaptionskip}{0.2cm} 
\setlength{\belowcaptionskip}{-0.15cm}
\centering
\tablestyle{3pt}{1.01}
\scalebox{0.99}{
\footnotesize
\begin{tabularx}{\linewidth}{@{}l|cccc@{}}
\toprule
\textit{\textbf{Multi-stage.}}  & \multicolumn{1}{c|}{\textbf{Train Time}}                    & \textbf{GPU}    & \textbf{Val}  & \textbf{Test} \\ \midrule
CLIMS~\cite{53}                           & \multicolumn{1}{c|}{1068 mins}                                 & 18.0 G          & 70.4          & 70.0          \\
MCTformer+~\cite{22}                      & \multicolumn{1}{c|}{1496 mins}                                 & 18.0 G          & 74.0          & 73.6          \\ \midrule
\textit{\textbf{Single-stage.}} &                                                                &                 &               &               \\ \midrule
AFA~\cite{27}                             & \multicolumn{1}{c|}{554 mins}                                  & 19.0 G          & 66.0          & 66.3          \\
ToCo~\cite{19}                            & \multicolumn{1}{c|}{506 mins}                                  & 17.9 G          & 71.1          & 72.2          \\
DuPL~\cite{51}                            & \multicolumn{1}{c|}{508 mins}                                  & 14.9 G          & 73.3          & 72.8          \\
\rowcolor[HTML]{EFEFEF} 
\textbf{MoRe(Ours)}             & \multicolumn{1}{c|}{\cellcolor[HTML]{EFEFEF}\textbf{372 mins}} & \textbf{12.1 G} & \textbf{76.4} & \textbf{75.0} \\ \bottomrule
\end{tabularx}
}
\caption{Efficiency of MoRe compared to others. The experiment is conducted on PASCAL VOC with RTX 3090.}
\vspace{-.4em}
\label{tab.5}
\end{table}

In addition, Table~\ref{tab.4} reports the class-specific segmentation performance of MoRe compared to recent methods and introduces Confusion Ratio (FP/TP) to investigate the ability to tackle artifact issues. MoRe shows a significantly lower confusion ratio over other methods, such as potted plant ({\textcolor{teal}{$-34.0\%$}} to DuPL), boat({\textcolor{teal}{$-30.0\%$}}), and higher IoU on all selected categories. For the average performance on VOC val set, MoRe achieves $0.22$ confusion ratio, lower by at least {\textcolor{teal}{$-9.0\%$}} to recent SOTAs, which validates the competence of MoRe to reduce false positives from artifact issues. 

\paragraph{Difference with Over-smoothness}
Over-smoothness of ViT is notorious for incurring false positives in WSSS. However, the reported artifact issue in our work shows different properties. To investigate it, the PTC loss~\cite{19} is incorporated in our ViT baseline, which is effective in tackling over-smoothness. As shown in Figure~\ref{fig3} (c), although LAM is regularized by PTC, artifacts still commonly exist and severally impair the quality. The reason lies in that over-smoothness arises from the high similarity among patch tokens while the artifact is derived from the improper relation between class-patch tokens. Previous methods barely pay efforts to regularize the class-patch attention, which impedes the development of LAM. By noticing this, MoRe regularizes such attention with GCR and LIR, which significantly improves the quality of LAM, as shown in Figure~\ref{fig3} (g,h).

\paragraph{Analysis of Training Efficiency}
Instead of designing sophisticated architectures, the proposed regularization modules can be seamlessly incorporated into ViT and effectively regularize the artifact. The training efficiency comparison is reported in Table \ref{tab.5}. The recent LAM-based method~\cite{22} cannot be implemented end-to-end, thus $1496$ minutes and $18.0$ GB GPU memory are taken to finish the WSSS workflow. In contrast, MoRe works in a single-stage manner and only needs $372$ minutes and $12.1$ GB GPU memory to finish the workflow, which significantly outperforms both multi-stage and single-stage counterparts.

\paragraph{Analysis of Hyper-parameters}
The analysis of important hyper-parameters in GCR and LIR, such as \textit{top-$K$, uncertain selection threshold, searching kernel size, etc.}, is specifically discussed in \textit{Appendix}.

\begin{figure}[t]
    \centering
    \includegraphics[width=1.0\linewidth]{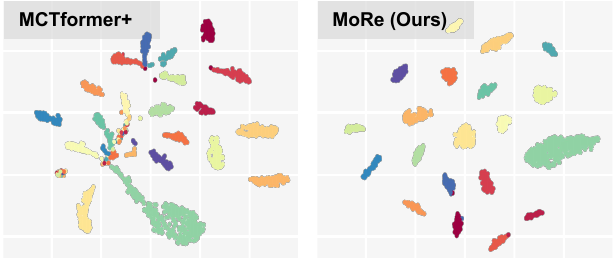}
    \vspace{-1.5em}
    \caption{Multi-class token representation between MCTformer+ and MoRe on VOC train set, visualized with t-SNE.}
    \label{fig5}
    \vspace{-1.2em}
\end{figure}

\paragraph{Analysis of Category Representation}
Previous LAM-based methods perform in-distinctive category representation, as shown in Figure~\ref{fig3} (b,c,d). To further investigate the role of the proposed regularization modules, we extract the multi-class tokens from the last layer of ViT encoder and visualize the representation with t-SNE~\cite{60} on VOC train set, as shown in Figure~\ref{fig5}. Although the LAM-based SOTA~\cite{22} intends to separate class tokens by minimizing their similarity, it is found that regularizing class tokens alone is insufficient to model separated feature space. In contrast, MoRe enhances the representation with a powerful topological graph and promotes the correlation between class tokens and related patch semantics, which demonstrates the distinguished class representation. Considering LAM is directly generated from the class-patch attention, the comparison supports the better pseudo mask performance in Table~\ref{tab.4}, Figure~\ref{fig3}, and verifies that MoRe builds a more robust class representation. 

In addition, when constrained to class tokens alone, the competitor struggles with the incompleteness and artifact issues as well, as shown in Figure~\ref{fig3} (d). Although it intends to refine LAM with CAM, the fusion strategy with a simple multiplying operation fails to improve the representation of categories and is still incompetent to cover the objects adequately, as shown in Figure~\ref{fig3} (e,f). In contrast, MoRe mines class-patch token relations from CAM and leverages it to supervise the attention, which further enhances the representation of categories and benefits the quality of both LAM and CAM, as shown in Figure~\ref{fig3} (g,h) and Table~\ref{tab1}, respectively. 

\section{Conclusion}
This work identifies and addresses the artifact issue when generating LAM from class-patch attention. We find that more regularization is needed to constrain the improper attention and propose MoRe to regularize it. Specifically, we propose the Graph Category Representation (GCR) module to model the class-patch attention as a directed graph, which implicitly excludes irrelevant artifact patches and aggregates useful information into class tokens. Then the Localization-informed Regularization (LIR) is proposed to explicitly promote the correlation of class and related patch tokens, which further helps suppress the artifacts and guarantees uncertain regions being activated. Extensive experiments and analysis are conducted on PASCAL VOC and MS COCO datasets, validating the efficiency of MoRe in tackling the artifact issue and significantly improving the performance of WSSS. 

\section{Acknowledgments}
This work was supported by the National Natural Science Foundation of China under Grant No.82372097, Shanghai Sailing Program under Grant 22YF1409300, International Science and Technology Cooperation Program under the 2023 Shanghai Action Plan for Science under Grant 23410710400, Taishan Scholars Program under Grant NO.tsqn202408245.

\bibliography{aaai25}
\end{document}